\documentclass{osa-article}

\journal{oe}

\articletype{Research Article}

\usepackage{lineno}

\begin{document}

\title{A Simple Self-calibration Method for The Internal Time Synchronization of MEMS LiDAR}

\author{Yu Zhang,\authormark{1} Xiaoguang Di,\authormark{2} Shiyu Yan,\authormark{1} Bin Zhang,\authormark{1} Baoling Qi,\authormark{1} and Chunhui Wang\authormark{1,*}}

\address{\authormark{1}National Key Laboratory of Tunable Laser Technology, Harbin Institute of Technology, Harbin, 150001, Heilongjiang, China,\\
\authormark{2}ontrol and Simulation Center,  Harbin Institute of Technology, Harbin, 150001, Heilongjiang, China}

\email{\authormark{*}wangch\_hit@163.com} 



\begin{abstract}
This paper proposes a simple self-calibration method for the internal time synchronization of MEMS(Micro-electromechanical systems) LiDAR during research and development. Firstly, we introduced the problem of internal time misalignment in MEMS lidar. Then, a robust Minimum Vertical Gradient(MVG) prior is proposed to calibrate the time difference between the laser and MEMS mirror, which can be calculated automatically without any artificial participation or specially designed cooperation target. Finally, actual experiments on MEMS LiDARs are implemented to demonstrate the effectiveness of the proposed method. It should be noted that the calibration can be implemented in a simple laboratory environment without any ranging equipment and artificial participation, which greatly accelerate the progress of research and development in practical applications.
\end{abstract}
\section{Introduction}
\label{sec1}
For most sensors, the calibrations of the intrinsic and extrinsic parameters are always necessary. With the development of camera and LiDAR technology, lots of calibration methods have been proposed, such as \cite{ZhengyouZhang2000}, \cite{8665256}, \cite{zuo2019lic}, \cite{zuo2020lic}, etc. However, most of these studies focus on improving the accuracy and convenience of calibration during the application stage, and few studies have reported methods to accelerate and facilitate the calibration process during research and development. 
\par In a MEMS LiDAR system, we can define an entire scan period as one frame, then ideally, the laser and MEMS mirror should have the same frame frequency and the same start time and end time at every frame. However, the time differences between the laser and MEMS mirror in LiDAR exist in every frame's start and end time, even in each row of a frame of data. 
\par
\begin{figure}[htbp]
\centering
\includegraphics [width=1\linewidth]{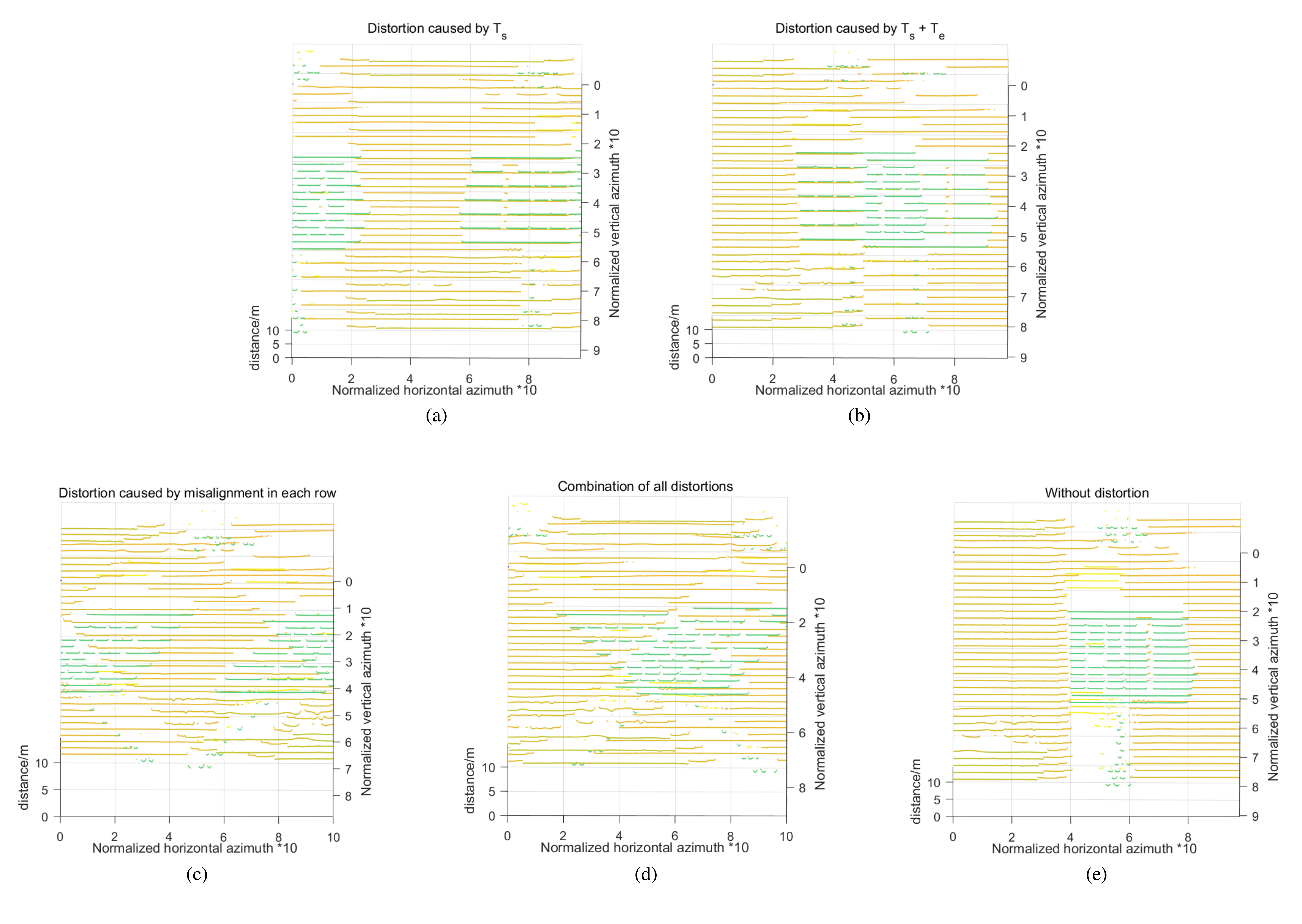}
\caption{The distortions caused by the time differences between the MEMS mirror and the laser. (a) Distortion caused by $T_s$. (b) Distortion caused by $T_s + T_e$. (c)Distortion caused by misalignment in each row. (d) Combination of all distortions. (e)Without distortion.}
\label{Fig:fig_distoration}
\end{figure}

 Firstly, MEMS mirror and laser are two different subsystems. Although they can be triggered synchronously, there is often a specific time difference due to the different signal distances and device response time. That leads to a time difference $T_s$ between the MEMS mirror and laser at the start of each frame, which will lead to obvious distortion of the point cloud generated by MEMS LiDAR, as shown in Fig. \ref{Fig:fig_distoration} (a).
\par At the same time, during the design, we first specify how many laser pulses a frame contains, and then formulate matching MEMS control signals. However, in the continuous scanning state, the actual response of the MEMS mirror is different from the theoretical value, which means the two independent control systems, MEMS mirror and laser, have different frame frequencies. That produces a time difference $T_e$ at the end of each frame, which will cause cumulative distortion of the generated point cloud, as shown in Fig. \ref{Fig:fig_distoration}(b). 
Also, there is a misalignment between laser and MEMS mirror in each row of one frame data. For example, at design time, MEMS scanned from one side to the other with 450 laser pulses. However, more or fewer pulses will be due to the response differences when the MEMS mirror scans a row. The misalignment will cause severe distortion of the generated point cloud, as shown in Fig. \ref{Fig:fig_distoration}(c). These time differences usually occur at the same time during the research and development process, as shown in Fig. \ref{Fig:fig_distoration}(d).
\par During the research and development process, we must constantly adjust the device to match the MEMS mirror and the laser, which is quite tedious and time-consuming. In this paper, we provide a Minimum Vertical Gradient(MVG) prior which can be used to calibrate those time differences in MEMS LiDAR instead of artificial alignment like \cite{Li2017mems}. The MVG prior greatly speeds up our debugging speed in reality.

\par Our contributions can be summarized as:
\begin{itemize}
\item We propose a Minimum Vertical Gradient(MVG) prior for MEMS LiDAR, which is robust to ranging error and can be used to automatically calibrate the distortion of point cloud caused by the internal time difference between the laser and MEMS mirror. 
\item The MVG prior is robust to ranging and azimuth error and can significantly accelerate research and development in practical applications. 
\item The proposed method has been tested on real MEMS LiDARs designed by us.
\end{itemize}
\par The remainder of the paper is organized as follows. In section \ref{sec2}, we explained the internal time synchronization problem in MEMS LiDAR in detail. In section \ref{sec3}, the MVG prior and the calibration method are stated. The actual experiments on MEMS LiDARs are implemented to demonstrate the effectiveness of the proposed method in section \ref{sec4}. Section \ref{sec5} concludes this paper and discusses the future research directions.
\section{MEMS LiDAR and Internal Time Synchronization}
\label{sec2}
\par
\begin{figure}[htbp]
\centering
\includegraphics [width=0.9\linewidth]{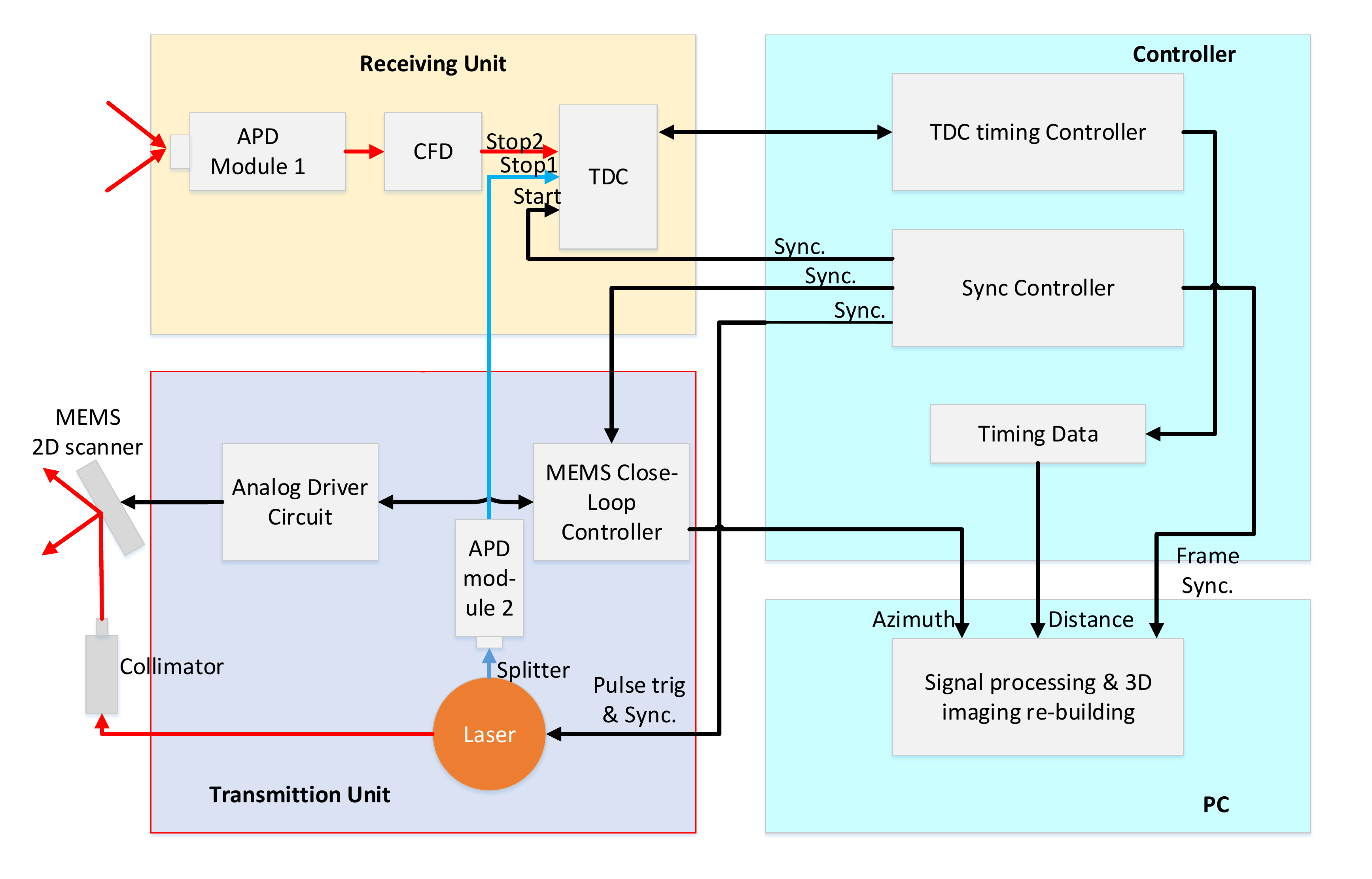}
\caption{ A typical structure of MEMS scanning LiDAR. The blue line represents the reference beam, the red line represents the measurement beam. The abbreviated information in the figure are: Avalanche PhotoDiode(APD), Constant Fraction Discrimination (CFD), Time to Digital Convert(TDC)}
\label{Fig:structure}
\end{figure}

 Typically, there are two kinds of LiDAR configuration, one is bi-static configuration, and the other is mono-static. The main difference between those two kinds of lidar is the number of antennas \cite{richmond2010direct}. Although the mono-configuration LiDAR has only one antenna to transmit and receive the pulse, which can save more space in LiDAR, it needs a more complex design than the bi-static one. Fig.\ref{Fig:structure} shows the structure of a bi-static configuration MEMS LiDAR, and the LiDAR equipment used in our experiment also adopts this structure. It can be seen that the MEMS LiDAR contains a two-dimensional rotating MEMS mirror and a laser transmitter. Through the MEMS mirror, the laser pulse can be deflected to different azimuths to realize the measurement of three-dimensional information.
\par The output information of MEMS LiDAR usually includes azimuths and measured distances. The measured distances are calculated through the TOF(Time Of Flight) method, which can be expressed as follows:
\begin{equation}
\label{eqn1}
R=c*\frac{t_{e}-t_{t}}{2}
\end{equation}
where $t_{e}$ and $t_{t}$ represent the actual time of the echo pulse and transmitted pulse, respectively, $c$ represents the speed of light ($c=30\  cm/ns $ for all the equations in this paper). The azimuths information can be calculated through the predetermined MEMS scanning path \cite{luodong2019}. We can regard azimuths and ranging distances as two periodic sequences, and one period contains one frame of data. The premise of the internal parameter correction \cite{garcia2020geometric,wang2019ukf} in the application phase is that these two sequences correspond accurately. However, during research and development, those two sequences often do not correspond to each other in the start and end of every period, and even period time.

\par
\begin{figure}[htbp]
\centering
\includegraphics [width=1\linewidth]{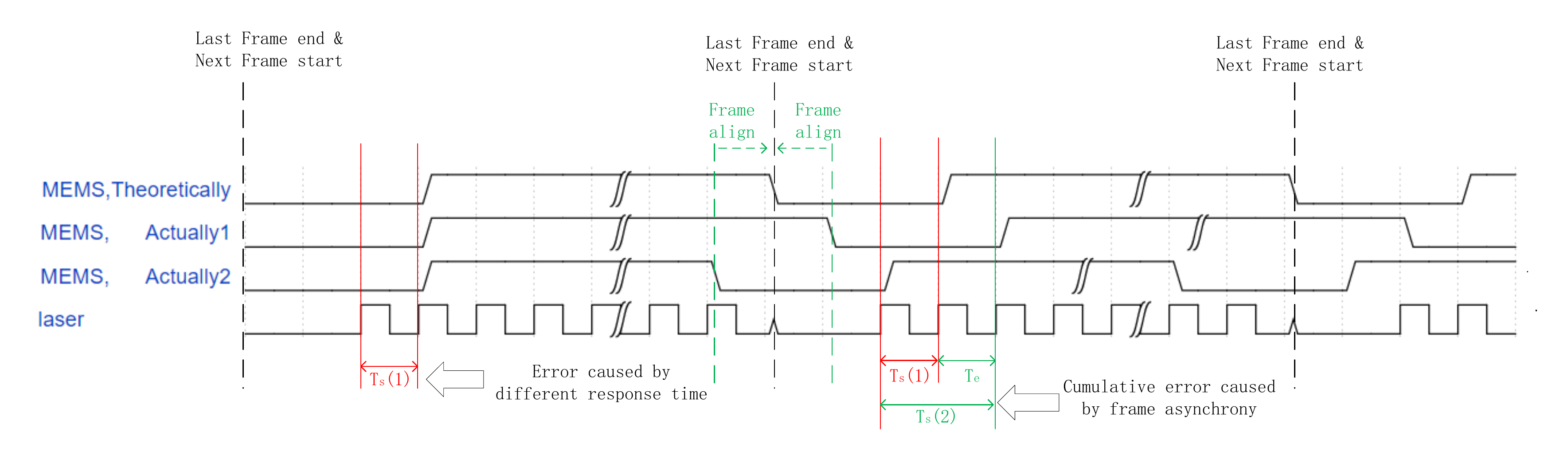}
\caption{Schematic diagram of the internal timing of the LiDAR. $T_s$ represents the start time difference between MEMS mirror and laser in one frame. When the frame frequencies of MEMS mirror and laser are different, $T_s$ will continue to change over time.}
\label{Fig:time}
\end{figure}

 For example, as shown in Fig.\ref{Fig:structure} and Fig.\ref{Fig:time}, the laser and MEMS mirror can share the same synchronization(sync.) signal, but in fact, they have different response time, which will produce a time difference $T_s$ between the MEMS mirror and laser at the scanning start of every frame(Usually, MEMS mirror requires a longer response time). $T_s$ will produce a mismatch between the azimuth and the measured distance, which will produce serious distortion in the point cloud (e.g., Fig.\ref{Fig:fig_distoration}(a)). In mechanical LiDAR, there are also time differences between the motor and the lasers. However, it will only cause the yaw azimuth shift of the point cloud, which can be easily solved by external parameters calibration with known objects. As far as we know, there is no previous research focusing on calibrating this time difference automatically. In \cite{Li2017mems}, an artificial time alignment method is provided, that is, artificially match the received distance signals and pre-designed azimuth through moving the distance signals one by one and observe whether the imaging results match the scene. However, this method is time-consuming, and the time difference changes a lot when the scanning path or angle changes. 
\par Also, during LiDAR scanning, the MEMS mirror is in continuous motion, and the response of the MEMS mirror is not entirely consistent with the control signals. Then, the actual value of the frame frequency is different from the theoretical value, which will cause the time difference $T_e$ at the end of the frame. As shown in Fig.\ref{Fig:time}, $T_e$ will accumulate to the $T_s$ of the next frame and make the distortions in next frame changing all the time. 
Typically, we must constantly adjust the MEMS control signals to make the MEMS mirror and laser have the same frame frequency. However, this is usually very difficult because we can only judge by comparing the actual scene and the point cloud image manually. Also, the different frame frequencies mean that the number of given azimuth and ranging information is not aligned. This further exacerbates the difficulty of the manual adjustment method. We need to manually align the start and end of each frame and each row of data to determine whether the MEMS mirror and laser are aligned accurately. In practice, we must reduce the scanning frame rate to ensure the distortion is not severe enough. Otherwise, the manual correction will be challenging to achieve. However, the distortions under different scanning frame rates are also different, so we cannot use the correction results at low scanning frame rates to replace high-speed conditions.

\section{Methods}
\label{sec3}
\subsection{Self-calibration of time difference between laser and MEMS mirror}
To avoid the complex artificial procedure during calibration, and inspired by the total variation (TV) loss \cite{rudin1992nonlinear} which is often used in low-level computer vision tasks, we propose a Minimum Vertical Gradient(MVG) prior to automatically calibrate the distorted point cloud. The TV loss is usually used to restrain noises, and treats the gradient as the total variation. It was proposed that the noisy images have higher total variation than clear images. The TV loss $J$ in an image can be expressed as follows:

\begin{equation}
\label{eqn2}
J_{TV}=\sum_{i=1}^{n} \sqrt{\left( \frac{\partial u_{i}}{\partial x}\right )^{2}+\left( \frac{\partial u_{i}}{\partial y}\right )^{2}} 
\end{equation}
where, $\frac{\partial u_{i}}{\partial x}$ and $\frac{\partial u_{i}}{\partial y}$ represent the horizontal and vertical gradient of pixel $i$ in image. Different from the TV loss, in the MVG prior, only the vertical gradients are used. And the vertical gradient is defined on a Polar coordinate system $ \left(\theta,\phi,R\right)$ , where $\theta$,$\phi$,$R$ represent the horizontal azimuth, vertical azimuth, and ranging distance, respectively.
The relationship between $\left(\theta,\phi,R\right)$ and the Cartesian coordinate system $\left(x,y,z\right)$ can be expressed as Equation \ref{eqn3}:
\begin{align}
\label{eqn3}
z=&\frac{R}{\sqrt{1+tan^{2}(\theta)+tan^{2}(\phi)}} \nonumber\\
y=& z*tan(\phi) \\
x=& z*tan(\theta) \nonumber
\end{align}
\par
\begin{figure}[htbp]
\centering
\includegraphics [width=0.7\linewidth]{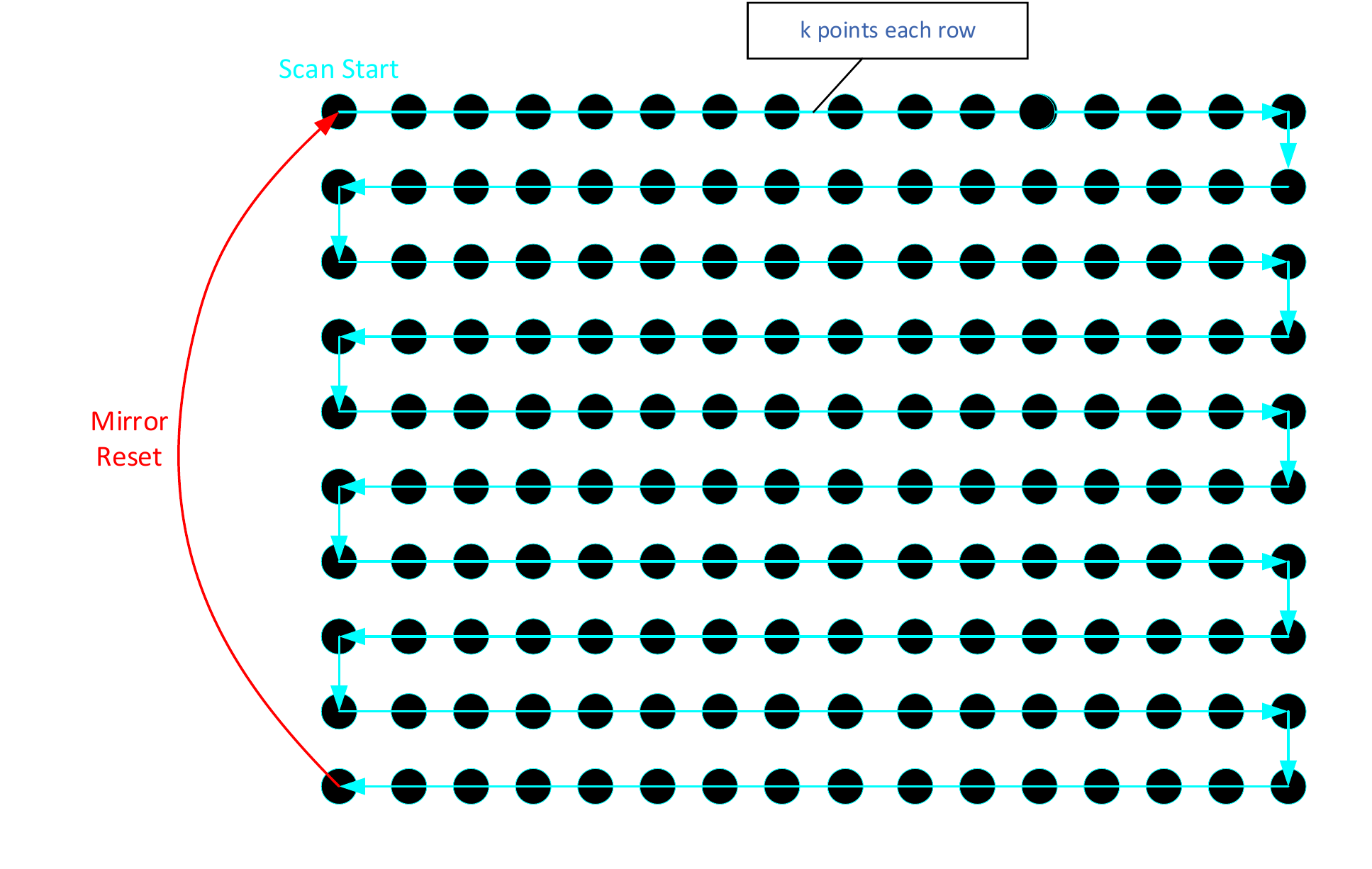}
\caption{MEMS scanning path. During the scanning process, in each frame, the MEMS mirror needs to be started from the same position, then it is needed to reset the MEMS position every frame.}
\label{Fig:raster}
\end{figure}

 When the MEMS mirror scans raster like Fig.\ref{Fig:raster}, the goal is to make the sequences of the measured distances and azimuths have the same starting, turning, and ending points, and the misalignment at those points will lead to the three time differences between laser and MEMS mirror, respectively. 
The two sequences after registration can be expressed as 
\begin{equation}
\label{eqn4}
R_{i+m}(\theta_{i}^{k},\phi_{i}^{k})
\end{equation}
where $m$ represents the num of pulses in the time difference $T_{s}$ at the start of every frame(e.g., $T_{s}$ in Fig. \ref{Fig:time}). $k$ represents the number of pulses corresponding to one row of raster that the MEMS mirror scans. $i$ represents the $i^{th}$ point in sequences.

Then the MVG prior can be expressed as follows:

\begin{equation}
\label{eqn5}
J_{MVG}=\underset{m,k}{min}\sum_{i=1}^{n-m} | \frac{\partial R_{i+m}(\theta_{i}^{k},\phi_{i}^{k})}{\partial \phi_{i}^{k} }| 
\end{equation}

\par From Equation \ref{eqn5}, we can get $m$ and $k$ for each frame of data. Then $T_s$ can be calculated through the folloing equation:
\begin{equation}
\label{eqn5_1}
 T_s = \Delta T * m
\end{equation}
where $\Delta T$ represents the time interval between two pulses. When the MEMS control signals are not modified, the value of $k$ is stable. However, it can be seen in Fig. \ref{Fig:time}, due to the different frame frequency between MEMS mirror and laser, $m$ will change over time. Therefore, we can denote $m$ as $m(t_{j})$, where $t_{j}$ represents the $t_{j}$ frame of ranging data. Also, in Fig. \ref{Fig:time}, it can be seen that the time differences $T_{e}$ caused by frame asynchrony will accumulate to $m(t_{j})$, so it can be calculated from 
\begin{equation}
\label{eqn6}
T_{e} = \frac{\Delta T}{n-1} \sum_{j=1}^{n-1}| m(t_{j+1})-m(t_{j})|
\end{equation}
\par After getting $T_{e}$, we can easily modify the control signals of the MEMS mirror in the position reset phase(e.g., the red part of Fig.\ref{Fig:raster}).

\section{Experiments}
\label{sec4}
To evaluate the effectiveness and accuracy of the proposed method, we verify the method on MEMS LiDARs designed and built by ourselves. A fiber pulse laser with a pulse width of $4 ns$ is adopted, and the transmitted pulse is split into measurement and reference beam with a splitting ratio of $95/5$. 

\begin{figure}[htbp]
\centering
\includegraphics [width=0.9\linewidth]{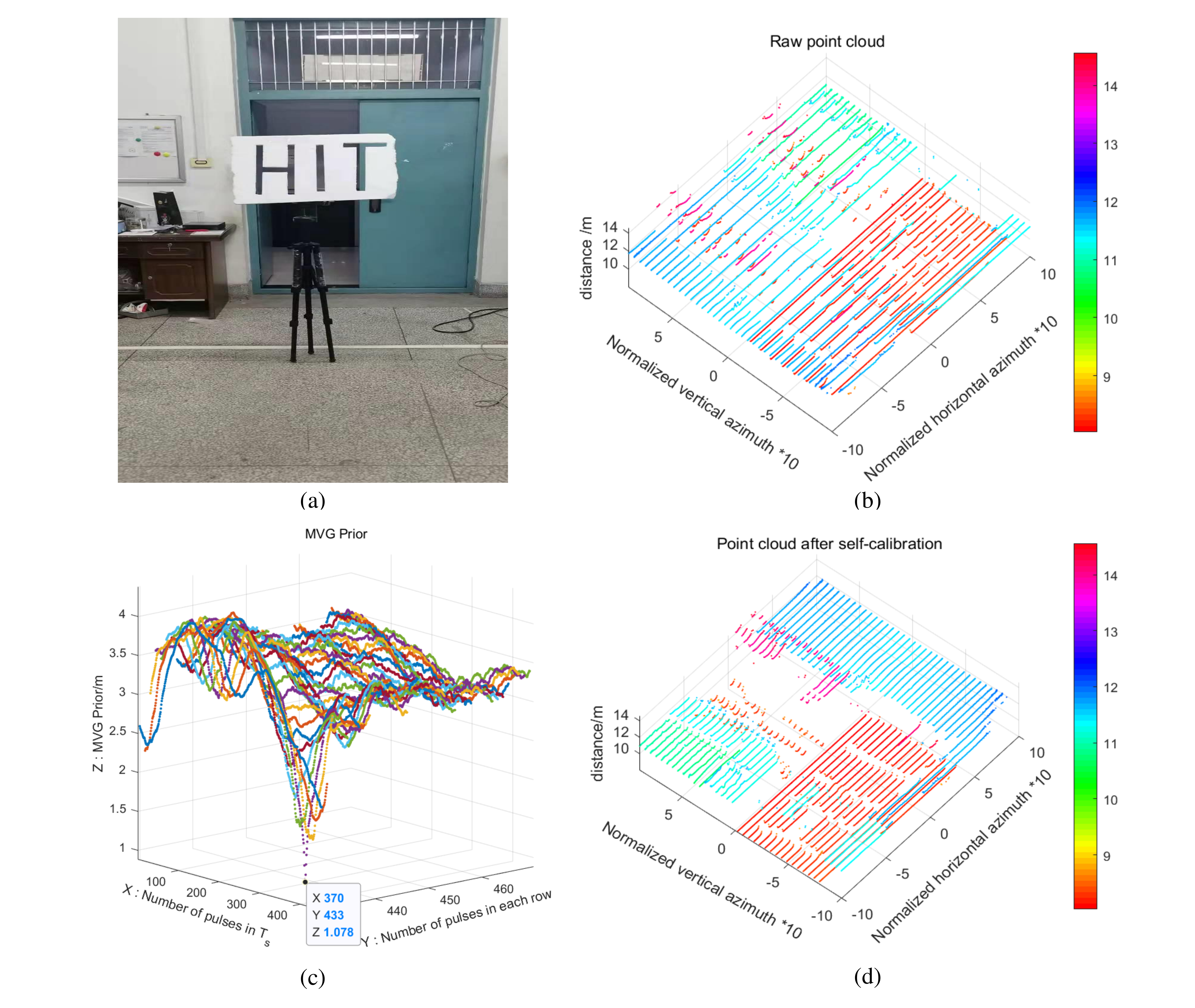}
\caption{The calibration result with MVG prior. (a) The image of the scanned scene. (b) Raw point cloud collected by LiDAR. (c) The MVG prior, X-axis represents the number of pulses in $T_{s}$, Y-axis represents the real number of pulses in each row of the data. (d) the calibration result.}
\label{Fig:fig_mvg}
\end{figure}

\begin{figure}[htbp]
\centering
\includegraphics [width=0.7\linewidth]{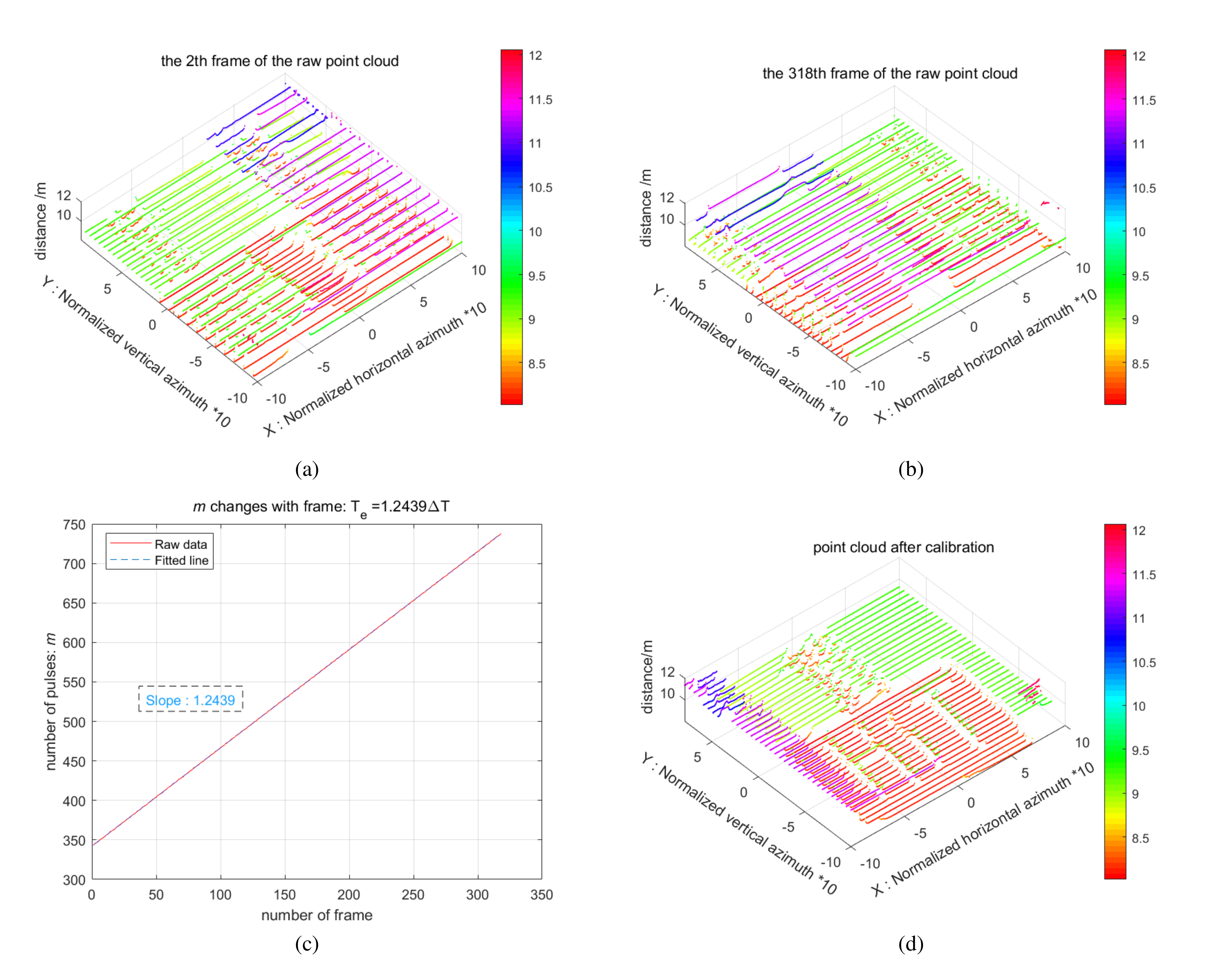}
\caption{The influence and calibartion of $T_{e}$. (a) The 2th frame of raw point cloud data. (b) The 318th frame of raw point cloud data. (c) m in $T_{s}$ changes with the frame cause by $T_{e}$. (d) The point cloud after calibration}
\label{Fig:fig_mvg_m_frame}
\end{figure}

\par As shown in Fig.\ref{Fig:fig_mvg}(b), the raw point cloud of LiDAR is highly distorted, and it is hard to identify the objects in the scene. However, with the MVG prior, as shown in Fig.\ref{Fig:fig_mvg}(c) and (d), we can easily get the numbers of pulses in $T_{s}$ and each row of the data. If with the manual-based method, we may need to proofread thousands of possible calibrated point clouds for one frame of data.
\par Fig.\ref{Fig:fig_mvg_m_frame} shows the influence and calibartion of $T_{e}$. As shown in Fig.\ref{Fig:fig_mvg_m_frame}(a) and (b), due to the existence of $T_{e}$, the distortion of the original point cloud data has been changing. However, with the MVG prior we can easily estimate the $T_e$ and increase the accuracy of $T_e$ to sub-pulse(e.g., Fig.\ref{Fig:fig_mvg_m_frame}(c)).
\par
\begin{figure}[htbp]
\centering
\includegraphics [width=0.7\linewidth]{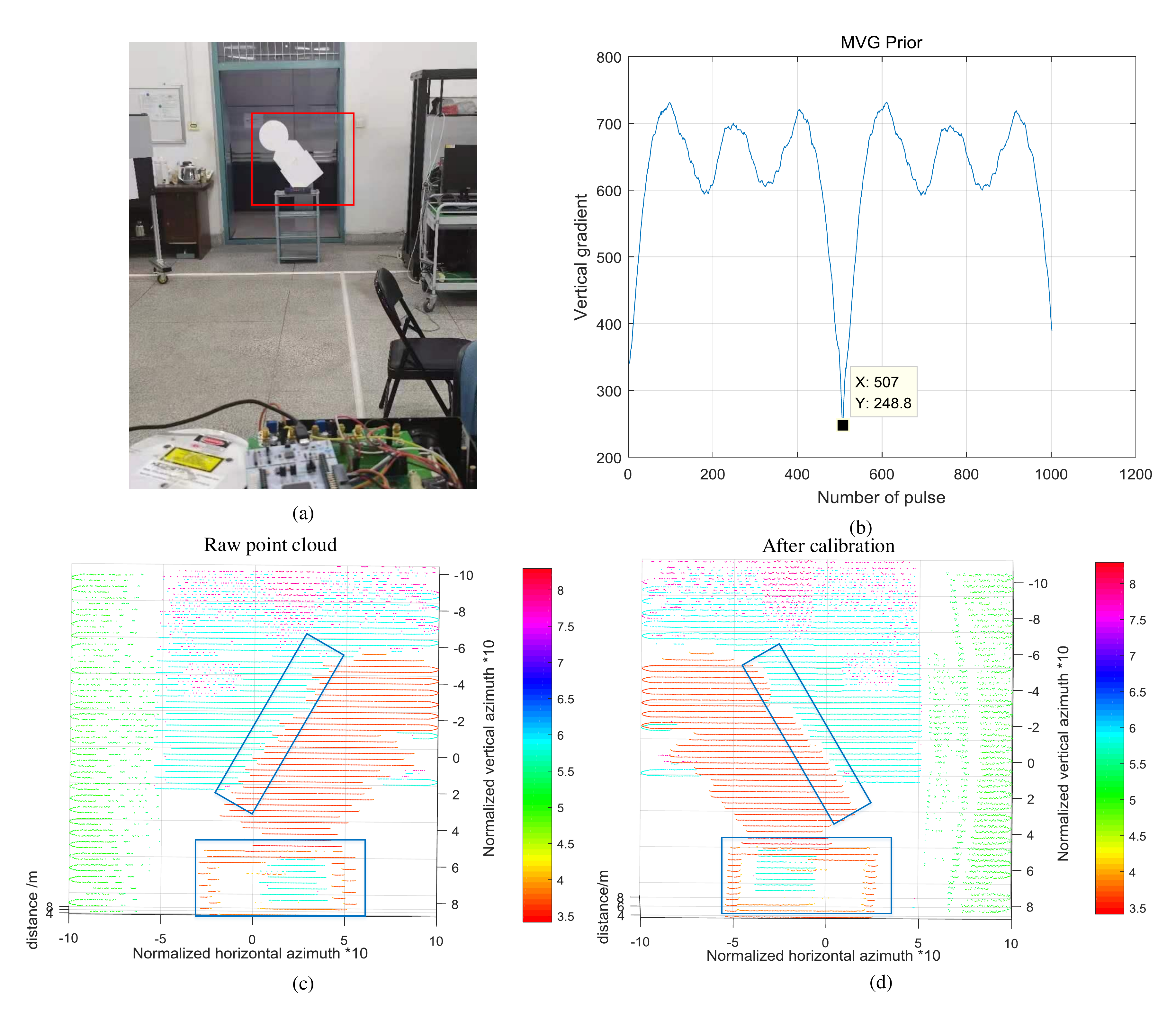}
\caption{The self-calibration of time difference between laser and MEMS mirror. (a) Scenes scanned by lidar. (b) The MVG prior. (c) Before calibration. (d) After calibration. }
\label{Fig:fig_human}
\end{figure}

\begin{figure}[htbp]
\centering
\includegraphics [width=0.7\linewidth]{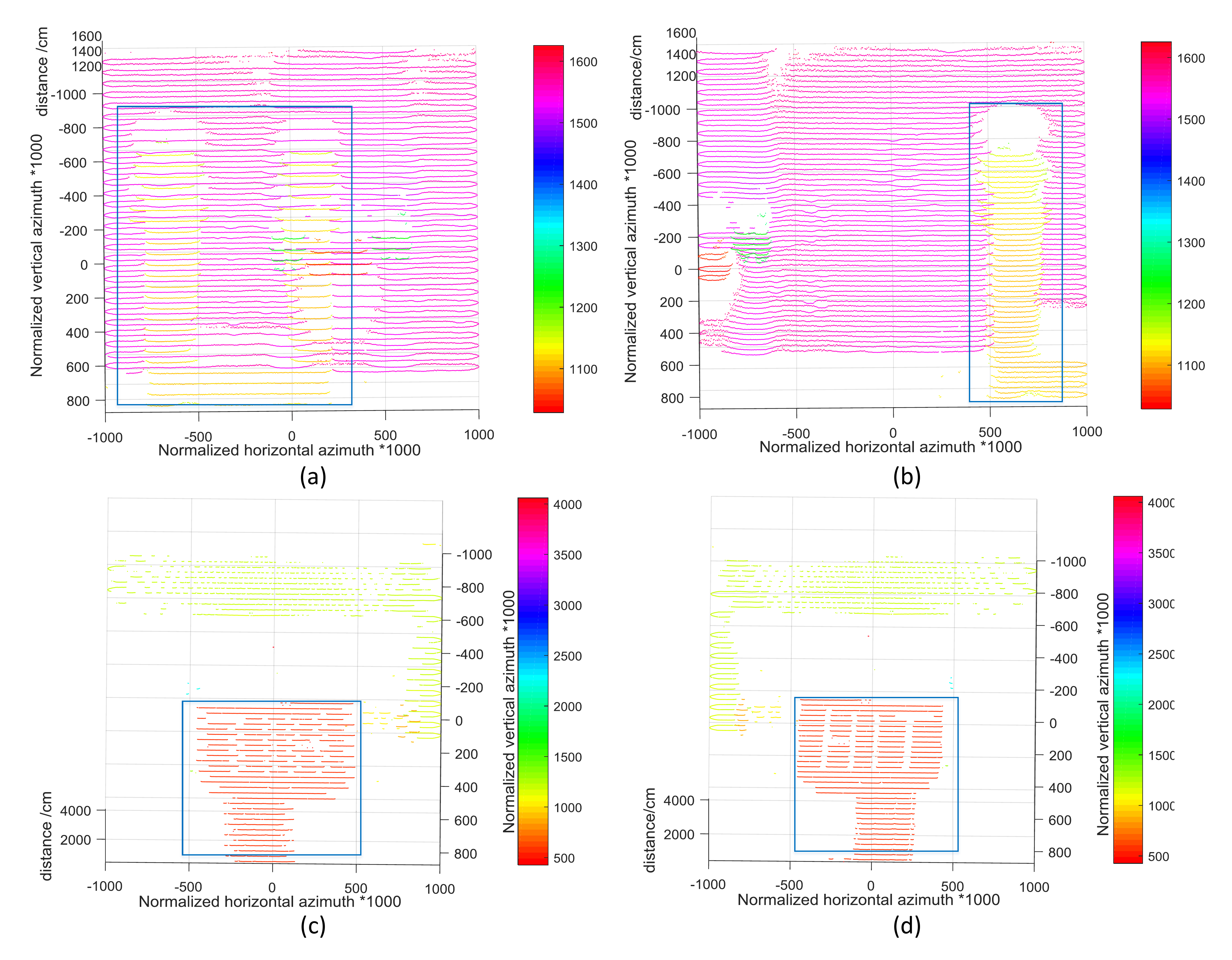}
\caption{The self-calibration of time difference between laser and MEMS mirror. (a) Result of smaller scanning angle and different scanning angle has time different between laser and MEMS mirror. (b) Calibration result of (a). (c) Result of longer scanning distance. (d) Calibration result of (c). }
\label{Fig:fig_other}
\end{figure}

 To evaluate the robustness of the MVG prior, we also test the method in different devices, with different object poses and distances, and with different scanning angles of the MEMS mirror. All the self-calibration results are in line with the real scene. Fig.\ref{Fig:fig_human} and Fig.\ref{Fig:fig_other} are tested on an early designed device. In this device, to avoid serious distortion in the case of high frame rates, the MEMS mirror scans at a very low speed(only $1.5$ frames per second), and the only time difference between the laser and MEMS mirror is $T_s$. Fig.\ref{Fig:fig_human} shows the MVG prior with the tilted target, and Fig.\ref{Fig:fig_other} shows the MVG prior when we change the scanning angle and target's distance. It can be seen that the MVG prior works well with those situations. Also, it should be noted that after applying the MVG prior, the scanning frame rate of the newly designed device is very easily increased to 10.84 frames per second.

\section{Conclusion}
\label{sec5}
\par In this paper, to speed up research and development and reduce manual works, we propose a robust and straightforward automatic calibration method for the internal time synchronization of MEMS LiDAR. Compared with the manual calibration methods, our method is faster and more convenient. The MVG prior can greatly improve the speed of research and development of MEMS lidar with a high scanning frame rate and variable scanning angle. Our future work will explore how to further improve the scanning angle and detection range of MEMS LiDAR.
\begin{backmatter}
\bmsection{Funding}
National Natural Science Foundation of China (61775048); Shenzhen Fundamental Research Program (JCYJ2020109150808037); National Key Scientific Instrument and Equipment Development Projects of China (62027823).
\bmsection{Disclosures}
The authors declare no conflicts of interest.
\bmsection{Data Availability Statement}
Data underlying the results presented in this paper are not publicly available at this time but may be obtained from the authors upon reasonable request.
\end{backmatter}
\bibliography{sample}






\end{document}